\documentclass[twocolumn,conference]{IEEEtran}
\usepackage[T1]{fontenc}
\usepackage{array}
\usepackage{float}
\usepackage{graphicx}
\usepackage{setspace}
\usepackage[unicode=true,
 bookmarks=true,bookmarksnumbered=true,bookmarksopen=true,bookmarksopenlevel=1,
 breaklinks=false,pdfborder={0 0 0},backref=false,colorlinks=false]
 {hyperref}
\hypersetup{pdftitle={Your Title},
 pdfauthor={Your Name},
 pdfpagelayout=OneColumn, pdfnewwindow=true, pdfstartview=XYZ, plainpages=false}

\makeatletter

\newcommand*\LyXZeroWidthSpace{\hspace{0pt}}
\providecommand{\tabularnewline}{\\}

\usepackage[caption=false,font=footnotesize]{subfig}
\usepackage{color}

\makeatother

\begin{document}

\title{DeepCough: A Deep Convolutional Neural Network in A Wearable Cough
Detection System}

\author{Justice Amoh, {\em Student Member, IEEE} and Kofi Odame, {\em Member, IEEE} \\ 
Thayer School of Engineering \\ 
Dartmouth College, Hanover, NH 03755, USA \\ 
{\bf E-mail:} {\ justice.amoh.jr.th@dartmouth.edu}}
\maketitle
\begin{abstract}
In this paper, we present a system that employs a wearable acoustic
sensor and a deep convolutional neural network for detecting coughs.
We evaluate the performance of our system on 14 healthy volunteers
and compare it to that of other cough detection systems that have
been reported in the literature. Experimental results show that our
system achieves a classification sensitivity of 95.1\% and a specificity
of 99.5\%.
\end{abstract}

\section{Introduction}

Automated real-time cough detection could be valuable for diagnosis
and treatment of airway diseases. Since coughs are relatively rare
events, a cough detector has to have a very low false alarm rate in
order to be useful at all. On the other hand, such a system needs
a high enough sensitivity in order to detect the infrequent event
of a cough. 

Researchers have attempted to address these challenges by employing
complex hand-crafted or hand-tuned features \cite{Barry2006,Matos2007}.
Unfortunately, these features can be time-consuming to develop and
may not necessarily be optimal for cough detection. In this study,
we propose a deep neural network framework for learning good features
that yield better cough detection.

\section{System Overview}

The proposed system for cough detection involves a wearable sensor
for acquiring the user's respiratory or vocal sounds in real-time.
The sensor, shown in Figure 1, is attached to the chest via a medical-grade
foam adhesive. Once worn, it streams lung and abdominal auscultation
sounds to a computer or smartphone for further processing and classification
of events. 

In contrast to previous systems that relied on conventional condenser
microphones \cite{Birring2008,Larson2011}, ours is a more application-specific
sensor consisting of a piezoelectric transducer and additional signal
conditioning electronics that enhance acoustic events of interest.
In particular, the sensor amplifies respiratory sounds, attenuates
voiced speech and eliminates environmental sounds altogether. Also,
the contact piezo-transducer captures the additional vibrational energy
that coughs induce in the body, which helps to improve discriminability. 

\begin{figure}[tbh]
\begin{centering}
\includegraphics[width=1\columnwidth]{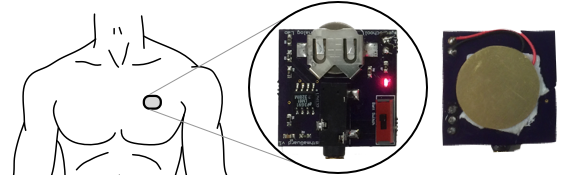}
\par\end{centering}

\protect\caption{An illustration of how the proposed sensor is worn. Front and back
images of sensor show the front-end electronics and the piezoelectric
transducer.}
\end{figure}

In the actual classification of coughs, previous works have thoroughly
explored hand-crafted features that are good for discriminating between
coughs and non-cough sounds. The Mel-Frequency Cepstral Coefficients
(MFCCs) and Linear Predictive Coding (LPC) are two popular examples.
While such features have proved effective in speech recognition and
have yielded impressive results in cough studies \cite{Barry2006,Matos2007},
they are not necessarily optimal for the specific task of cough detection.
Furthermore, those features have typically been used with conventional
condenser microphone signals and may not be appropriate for our contact
piezo sensor. So, our proposed system employs the deep convolutional
neural network framework to jointly learn good feature representations
and to train a robust classifier, suitable for our sensor and for
the cough classification task.

\begin{figure*}[tbh]
\centering{}\includegraphics[width=0.8\paperwidth]{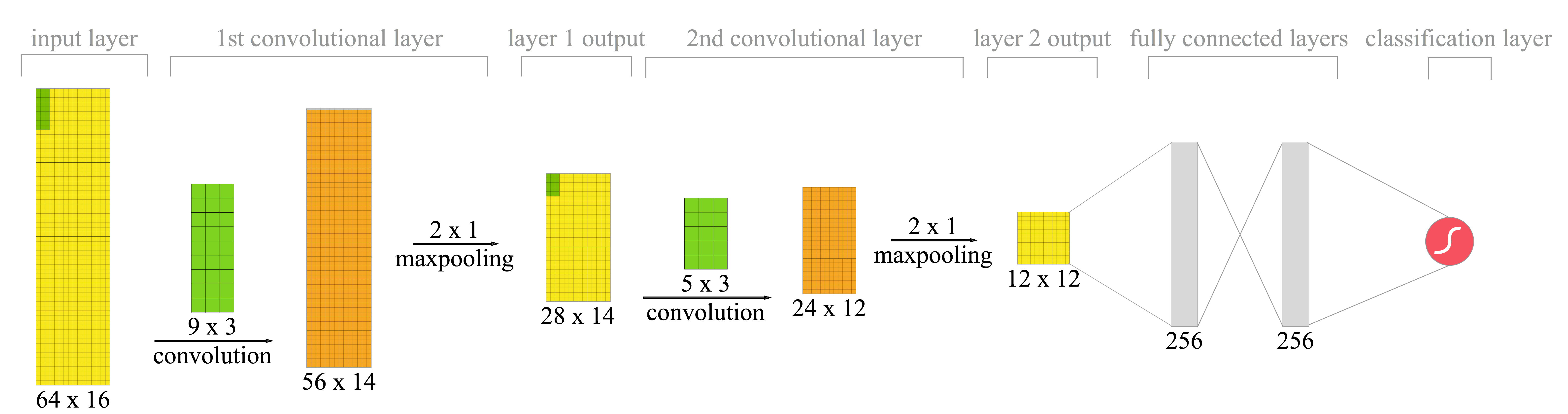}\protect\caption{An illustration of the architecture of our CNN. The input to the network
is a 64 ms STFT spectrogram. Network consists of 2 conv. layers, 2
fully connected layers and a softmax classification layer. Each conv.
layer has 16 filters, with rectified linear unit (ReLU) activations.
Filter and pooling sizes are chosen to encourage learning of patterns
across both temporal and spectral domains.}
\end{figure*}

\section{Cough Detection Details}

Our cough detection algorithm involves a preprocessing stage, followed
by a classifier. The preprocessing stage extracts some preliminary
features, and it also limits the amount of uninteresting data that
is admitted to the classifier; the classifier is a convolutional neural
network (CNN) that performs further feature extraction and labels
an audio event as either cough or not.

\subsection{Preprocessing}

During preprocessing, the stream of acoustic data is segmented into
frames which are each 4 ms long. To eliminate irrelevant data such
as silence and background noise, the preprocessor implements the frame
admission process suggested by \emph{Lu et al} \cite{Lu2009}. For
each 16-frame (64 ms) window, the RMS energy is calculated and compared
with a predetermined threshold. Windows with low energy are assumed
to be silence or ambient sound, and they are discarded. High energy
windows are \textquotedbl{}admitted\textquotedbl{} and undergo further
processing.

For each admitted window, a 128-bin Short Time Fourier Transform (STFT)
is performed to yield a 64$\times$16 spectral segment (i.e. spectrogram
of 64 ms data). These spectral segments are the inputs which are fed
to the CNN for classification.

\subsection{Convolutional Neural Network Architecture}

CNNs are multilayered perceptron (MLP) models, whose neurons mimic
the mammalian visual cortex in that they are sensitive to small overlapping
regions of the receptive field \cite{Ng}. Recently, CNNs have been
successfully applied to several computer vision tasks such as object
recognition, localization and tracking \cite{Krizhevsky2012,Donahue2014}.
CNNs have also been applied to other domains such as action recognition,
speech recognition, natural language processing and music genre classification
\cite{Hinton2012}. In our work, we implement a CNN to discriminate
between cough and speech sounds. 

Our network consists of five layers: 2 convolutional layers, 2 fully
connected layers and a softmax classification layer. Each convolutional
layer has 16 rectified linear units (ReLU). The first convolutional
layer takes in the 64$\times$16 spectral segments as inputs, and
has filters of size 9$\times$3. This is followed by a 2$\times$1
maxpooling layer. The second convolutional layer has filters of size
5$\times$3 and is also followed by a 2$\times$1 maxpooling layer.
Convolutions are performed with a stride of 1. The convolutional layers
are followed by 2 fully connected layers with 256 rectified linear
units each. Each fully connected layer also employs dropout regularization
(p=0.5) to reduce overfitting. Finally, the last layer takes the outputs
of the second fully connected layer and classifies the input as either
a cough or speech event using the softmax function. The network architecture
is illustrated in Figure 2.

We chose ReLU activations over the traditional tanh or sigmoid functions
because ReLU doesn't have the vanishing gradient problem and often
leads to a faster convergence \cite{Krizhevsky2012}. The convolutional
filter sizes are chosen to enable 2D convolutions: across both frequency
and temporal domains. Previous applications of convolutional networks
in audio sometimes convolved along either time or frequency axis \cite{Lee2009}.
For our application however, since we know both short-term temporal
and spectral patterns can be discriminative for cough and speech events,
we convolve along both dimensions. In addition, since our input segments
cover a relatively short time window (16 frames, 64 ms), we fix the
size of the filters along the time axis (at 3 frames). Pooling layers
downsample outputs of convolutions to make computations manageable
in subsequent layers. So just like in our filter sizing, we perform
no pooling along the time axis to avoid further reducing the rather
limited temporal resolution of segments.

\section{Experimental Methods}

\subsection{Data Collection}

To build and evaluate the proposed system, we created a database of
lung sound recordings from 14 healthy volunteers: 7 males and 7 females.
All subjects provided informed consent, and the experimental protocol
was approved by the Dartmouth College Institutional Review Board.
The piezo sensor was used to collect acoustic data as subjects were
guided through a series of procedures, including producing forced
bouts of coughs and reading some prompts out loud. Each subject produced
an average of 40 cough sounds, yielding a total of 627 cough examples
in our database. For the speech data, subjects read out 20 phonetically-balanced
prompts from the Harvard Sentences database \cite{Rothauser1969}.
Snippets were extracted from these sentence recordings, so that there
were equal numbers of speech and cough samples, and each sample was
of similar length. For an objective assessment of our piezo sensor,
we also used a professional digital recorder, an Olympus LS-12, to
simultaneously record all the speech and cough sounds. Both the piezo
sensor and the Olympus LS-12 recorder were sampled at a 44.1 kHz rate
and later down-sampled to 16 kHz.

\subsection{Network Training}

In training our neural network, we first split our database in two
parts: 70\% for building the model, and 30\% for testing. We further
split the model-building data into training and validation sets in
an 80:20 ratio. The training set is used to actually train the network.
The trained model is then run several times against the validation
set to find optimal model hyper-parameters (eg. learning rate, number
of filters, etc). Once all hyperparameters are found, the model is
retrained and run against the test set for the final evaluation. 

We augment our input data to introduce some translational invariance
in the learning. This is done by re-buffering the 16-frame spectral
segments from the same events to have a 4-frame overlap (25\%). Segments
at the edges are zero-padded as necessary. Our training database resolves
into 10,279 segments, with which we train our network. We also standardize
the entire training data across all components as is often done in
training deep neural networks. \LyXZeroWidthSpace{}

The convolutional network is trained using stochastic gradient descent,
with a learning rate of 0.001, batch size of 20 and a momentum of
0.9. Training converges after roughly 50 epochs, with a run time of
about one hour. Implementation of the convolutional network is done
using Lasagne \cite{sander_dieleman_2015_27878}, a Theano-based library
for training neural networks.

\subsection{Experiments}

To evaluate the performance of the proposed system, two experiments
are undertaken. The first investigates the hypothesis that the CNN
extracts better features for cough detection than the traditional
hand-crafted MFCC features. The second experiment compares the entire
end-to-end detection system with alternative approaches.

\subsubsection{Experiment 1}

To verify how effective the learned CNN features are for cough classification,
we compare against MFCC features. We extract 13 MFCC coefficients
from every 8 ms of training examples, with 4 ms overlap (50\%). This
yields an equivalent number of frames as in the STFT spectral segments
(16 per 64 ms) used in training the CNN. Since the classifcation layer
of the CNN is a softmax, a softmax function is also trained on these
MFCC features. This is to allow for a direct comparison of the representational
abilities of the MFCC and CNN with respect to our cough detection
task. We also train a Support Vector Machine (SVM) on the MFCC features
to observe the potential improvements a more complex classifier could
provide. In addition, we train a linear SVM on the raw STFT data to
serve as a reference bar for comparison with the CNN features.

\subsubsection{Experiment 2}

In the second experiment, we compare the two main aspects of our proposed
system with the conventional Hidden Markov Model approach. 

The duration of coughs in our database ranges from 250 ms to about
800 ms. To ensure our speech data is in a similar format, we split
each prompt recording into smaller segments with random durations
in the same range as those of the cough examples. In all, the average
duration of a test example is 320 ms, which makes for a comparable
test setup for both the CNN model and the HMM model. For each such
\textasciitilde{}320 ms window of acoustic data, the HMM model outputs
a single class prediction. On the other hand, the CNN model would
yield \textasciitilde{}4 class predictions, one for each 64ms segment.
Hence, to obtain a CNN prediction for the entire window, we average
over the classification scores (probabilities) for all segments in
the window. 

With the above testing framework, we first investigate how our system
compares with traditional MFCC-based HMM. An HMM with 10 states is
trained for each of the classes. The first and last states are non-emitting,
but all middle states have an emission probability distribution modeled
by a 13-dimensional mixture of Gaussians. For each training example,
13 MFCC coefficients are extracted for every 25 ms frame and the sequence
of these coefficient vectors is used to train the HMMs. At test time,
a similar feature vector sequence extracted from the test example
is fitted to both HMMs. The resulting log-likelihood values of both
fits determines whether the sound pertains to a cough or speech event.
This HMM configuration is fairly common in cough studies and in speech
recognition \cite{Matos2006}. 

Using the same model setup and testing framework, we also investigate
how the piezo sensor compares with conventional condenser microphones,
for cough detection. Since we have accompanying microphone recordings
for all collected sensor data, we train HMM and CNN models on microphone
data and compare its performance with that of the piezo sensor.

\section{Results \& Discussion}

Table 1 shows the results for experiment 1. First, we observe that
our CNN model performs much better (\textasciitilde{}10\% more) than
training an SVM on the raw STFT data. This agrees with the notion
that the CNN is indeed extracting features from the STFT, that are
useful for the classification task. Furthermore, based on this dataset,
the CNN model appears to outperform the MFCC with either the softmax
classifier or the SVM, suggesting that the CNN is a more effective
feature extractor for cough classification. An interesting observation
we note is that the MFCC+SVM model yields a specificity comparable
with that of the CNN model. Specificity, in this binary cough-speech
discrimination, refers to the model's accuracy in detecting speech
events. And so a possible explanation for why MFCC still manage to
yield a high specificity could be the fact that MFCCs are particularly
designed to mimic how we humans hear and is known to be very effective
for speech recognition and processing. A fairly able classifier like
the SVM can leverage this innate MFCC representational abilities for
speech for a better specificity. The STFT+SVM model performance supports
this idea of an MFCC speech advantage, seeing as it has comparable
sensitivity as the MFCC+SVM model, but fails to match its specificity. 

\begin{table}[htbp]
\protect\caption{Comparison of CNN and MFCC For Cough Classification }

\begin{onehalfspace}
\centering{}{\small{}}%
\begin{tabular}{|l|c|c|}
\hline 
\textbf{\footnotesize{}Model} & \textbf{\footnotesize{}Sensitivity} & \textbf{\footnotesize{}Specificity}\tabularnewline
\hline 
\hline 
{\scriptsize{}MFCC+SM} & {\footnotesize{}$87.5$} & {\footnotesize{}$86.2$}\tabularnewline
\hline 
{\scriptsize{}MFCC+SVM} & {\footnotesize{}$86.3$} & {\footnotesize{}$90.7$}\tabularnewline
\hline 
{\scriptsize{}STFT+SVM} & {\footnotesize{}$84.2$} & {\footnotesize{}$80.3$}\tabularnewline
\hline 
\textbf{\scriptsize{}STFT+CNN} & \textbf{\footnotesize{}94.0} & \textbf{\footnotesize{}91.7}\tabularnewline
\hline 
\end{tabular}\end{onehalfspace}
\end{table}

\begin{table*}[t]
\protect\caption{Comparison with Previous Cough Detection Works}

\centering{}%
\begin{tabular}{|c|c|c|c|c|c|}
\hline 
\textbf{\footnotesize{}Study} & \textbf{\footnotesize{}Model} & \textbf{\footnotesize{}Subjects} & \textbf{\footnotesize{}Decision Window (ms)} & \textbf{\footnotesize{}Sensitivity (\%)} & \textbf{\footnotesize{}Specificity (\%)}\tabularnewline
\hline 
\hline 
{\footnotesize{}HACC \cite{Barry2006}} & {\footnotesize{}LPC-PNN} & {\footnotesize{}15} & {\footnotesize{}1000} & {\footnotesize{}80.0} & {\footnotesize{}96.0}\tabularnewline
\hline 
{\footnotesize{}LCM \cite{Matos2007}} & {\footnotesize{}MFCC-HMM} & {\footnotesize{}19} & {\footnotesize{}10000} & {\footnotesize{}85.7} & {\footnotesize{}94.7}\tabularnewline
\hline 
{\footnotesize{}Swarnkar et al. \cite{Swarnkar2013}} & {\footnotesize{}NN} & {\footnotesize{}3} & {\footnotesize{}100} & {\footnotesize{}93.4} & {\footnotesize{}94.5}\tabularnewline
\hline 
{\small{}Larson et al. \cite{Larson2011}} & {\small{}PCA-Random Forest} & {\small{}17} & {\small{}150} & {\small{}92.0} & \textbf{\small{}99.5}\tabularnewline
\hline 
\textbf{\footnotesize{}DeepCough}{\footnotesize{} 1} & {\footnotesize{}CNN} & {\footnotesize{}14} & {\footnotesize{}64} & {\footnotesize{}94.0} & {\footnotesize{}91.7}\tabularnewline
\hline 
\textbf{\footnotesize{}DeepCough}{\footnotesize{} 2} & {\footnotesize{}CNN} & {\footnotesize{}14} & {\footnotesize{}320} & \textbf{\footnotesize{}95.1} & \textbf{\footnotesize{}99.5}\tabularnewline
\hline 
\end{tabular}
\end{table*}

The Receiver Operating Characteristic (ROC) curves in Figure 3 show
how our proposed system compares with the conventional microphone
and HMM approaches. Our CNN based system substantially outperforms
the HMM models with an AUC close to 1 in both piezo sensor and microphone
scenarios. 

\begin{figure}[H]
\begin{minipage}[t]{0.5\columnwidth}%
\begin{center}
\includegraphics[scale=0.2]{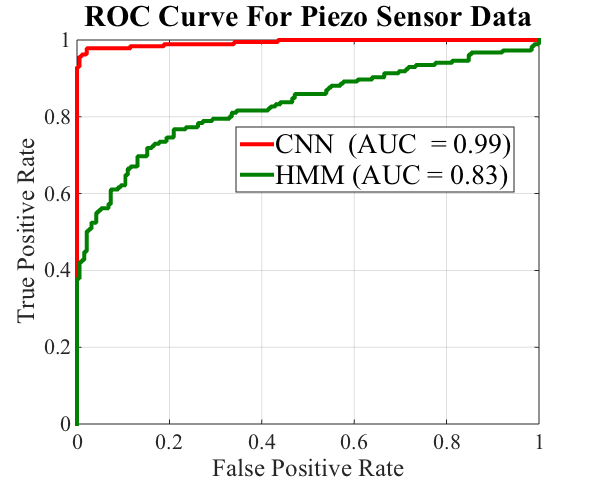}
\par\end{center}

\begin{center}
(a)
\par\end{center}%
\end{minipage}%
\begin{minipage}[t]{0.48\columnwidth}%
\begin{center}
\includegraphics[scale=0.2]{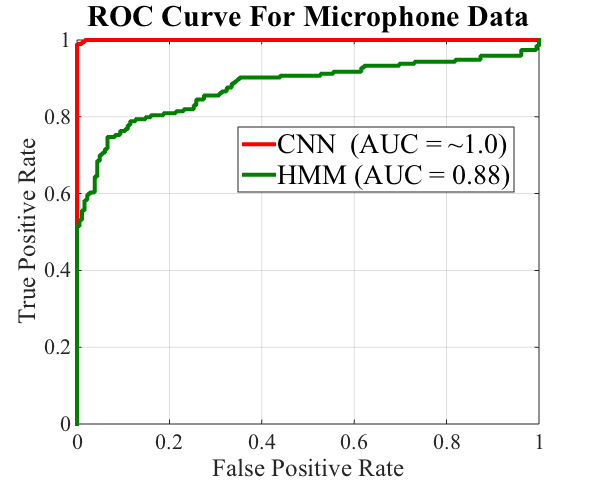}
\par\end{center}

\begin{center}
(b)
\par\end{center}%
\end{minipage}

\protect\caption{Receiver Operating Characteristic (ROC) curves for both CNN and HMM
models on (a) Piezoelectric sensor data and on (b) Microphone data.
Performance is similar across piezo and mic, although slightly better
for microphone.}
\end{figure}

In Table 2, we compare our system with other previous works in cough
detection. Based on our data, our model seems to outperform the others
with a sensitivity of 95.1\%. However, the reported specificity values
may not be directly comparable with our work since the other studies
consider silence and low energy events as false negatives too.

\section{Conclusions}

In this study, we proposed a system that employs a wearable acoustic
sensor and a deep convolutional neural network for detecting coughs.
We evaluated our model's ability to extract good features for the
custom sensor and for the cough detection task. We also show that
our convolutional network model outperforms previous works in the
literature. 

Future works will explore other neural network architectures more
suitable to time varying inputs such as Recurrent Neural Networks.
A more immediate follow up study would be an extended passive data
collection where subjects or patients can wear our sensors for hours
or days to capture un-forced coughs.

\section*{Acknowlegment}

This work is supported by the Neukom Institute for Computational Science
at Dartmouth College. We would like to thank Professor Lorenzo Toressani
(Computer Science Department, Dartmouth College) for discussions and
suggestions on deep learning methods and resources. 

\bibliographystyle{IEEEtran}
\bibliography{BiocasBib}

\begin{thebibliography}{10}
\providecommand{\url}[1]{#1}
\csname url@samestyle\endcsname
\providecommand{\newblock}{\relax}
\providecommand{\bibinfo}[2]{#2}
\providecommand{\BIBentrySTDinterwordspacing}{\spaceskip=0pt\relax}
\providecommand{\BIBentryALTinterwordstretchfactor}{4}
\providecommand{\BIBentryALTinterwordspacing}{\spaceskip=\fontdimen2\font plus
\BIBentryALTinterwordstretchfactor\fontdimen3\font minus
  \fontdimen4\font\relax}
\providecommand{\BIBforeignlanguage}[2]{{%
\expandafter\ifx\csname l@#1\endcsname\relax
\typeout{** WARNING: IEEEtran.bst: No hyphenation pattern has been}%
\typeout{** loaded for the language `#1'. Using the pattern for}%
\typeout{** the default language instead.}%
\else
\language=\csname l@#1\endcsname
\fi
#2}}
\providecommand{\BIBdecl}{\relax}
\BIBdecl

\bibitem{Barry2006}
S.~J. Barry, A.~D. Dane, A.~H. Morice, and A.~D. Walmsley, ``{The automatic
  recognition and counting of cough.}'' \emph{Cough (London, England)}, vol.~2,
  p.~8, Jan. 2006.

\bibitem{Matos2007}
S.~Matos, S.~S. Birring, I.~D. Pavord, and D.~H. Evans, ``{An automated system
  for 24-h monitoring of cough frequency: the leicester cough monitor.}''
  \emph{IEEE transactions on bio-medical engineering}, vol.~54, no.~8, pp.
  1472--9, Aug. 2007.

\bibitem{Birring2008}
S.~S. Birring, T.~Fleming, S.~Matos, a.~a. Raj, D.~H. Evans, and I.~D. Pavord,
  ``{The Leicester Cough Monitor: Preliminary validation of an automated cough
  detection system in chronic cough},'' \emph{European Respiratory Journal},
  vol.~31, no.~5, pp. 1013--1018, 2008.

\bibitem{Larson2011}
E.~C. Larson, T.~Lee, S.~Liu, M.~Rosenfeld, and S.~N. Patel, ``{Accurate and
  privacy preserving cough sensing using a low-cost microphone},''
  \emph{Proceedings of the 13th international conference on Ubiquitous
  computing - UbiComp '11}, p. 375, 2011.

\bibitem{Lu2009}
H.~Lu, W.~Pan, N.~Lane, T.~Choudhury, and A.~Campbell, ``{SoundSense: scalable
  sound sensing for people-centric applications on mobile phones},''
  \emph{Proceedings of the 7th international conference on Mobile systems,
  applications, and services}, pp. 165--178, 2009.

\bibitem{Ng}
\BIBentryALTinterwordspacing
A.~Ng, J.~Ngiam, and C.~Y. Foo, ``{Unsupervised Feature Learning and Deep
  Learning Tutorial}.'' [Online]. Available:
  \url{http://deeplearning.stanford.edu/tutorial/}
\BIBentrySTDinterwordspacing

\bibitem{Krizhevsky2012}
A.~Krizhevsky, I.~Sutskever, and G.~E. Hinton, ``{ImageNet Classification with
  Deep Convolutional Neural Networks},'' \emph{Advances In Neural Information
  Processing Systems}, pp. 1--9, 2012.

\bibitem{Donahue2014}
J.~Donahue, Y.~Jia, O.~Vinyals, J.~Hoffman, N.~Zhang, E.~Tzeng, and T.~Darrell,
  ``{DeCAF: A Deep Convolutional Activation Feature for Generic Visual
  Recognition},'' \emph{International Conference on Machine Learning}, vol.~32,
  pp. 647--655, 2014.

\bibitem{Hinton2012}
G.~Hinton, L.~Deng, D.~Yu, G.~E. Dahl, A.-r. Mohamed, N.~Jaitly, A.~Senior,
  V.~Vanhoucke, P.~Nguyen, T.~N. Sainath, and B.~Kingsbury, ``{Deep Neural
  Networks for Acoustic Modeling in Speech Recognition},'' \emph{Ieee Signal
  Processing Magazine}, no. November, pp. 82--97, 2012.

\bibitem{Lee2009}
H.~Lee, P.~Pham, Y.~Largman, and A.~Ng, ``{Unsupervised feature learning for
  audio classification using convolutional deep belief networks.}''
  \emph{Nips}, pp. 1--9, 2009.

\bibitem{Rothauser1969}
E.~H. Rothauser, W.~D. Chapman, N.~Guttman, H.~R. Silbiger, and J.~L. Sullivan,
  ``{IEEE recommended practice for speech quality measurements},'' \emph{IEEE
  Trans. Audio \ldots}, vol. AU-17, no. 297, pp. 225--246, 1969.

\bibitem{sander_dieleman_2015_27878}
\BIBentryALTinterwordspacing
S.~Dieleman, J.~Schl\"{u}ter, C.~Raffel \emph{et~al.}, ``{Lasagne: First
  release.}'' Aug. 2015. [Online]. Available:
  \url{http://dx.doi.org/10.5281/zenodo.27878}
\BIBentrySTDinterwordspacing

\bibitem{Matos2006}
S.~Matos, S.~Member, S.~S. Birring, I.~D. Pavord, D.~H. Evans, and S.~Member,
  ``{Detection of Cough Sounds in Continuous Audio Recordings Using Hidden
  Markov Models},'' vol.~53, no.~6, pp. 1078--1083, 2006.

\bibitem{Swarnkar2013}
V.~Swarnkar, U.~R. Abeyratne, S.~M. Ieee, Y.~Amrulloh, C.~Hukins, and
  R.~Triasih, ``{A Neural Network Based Algorithm for Automatic Identification
  of Cough Sounds},'' pp. 1764--1767, 2013.

\end{thebibliography}

\end{document}